\documentclass{article}

 \usepackage[dblblindworkshop, final]{neurips_2025}

\usepackage[utf8]{inputenc} 
\usepackage[T1]{fontenc}    
\usepackage{hyperref}       
\usepackage{url}            
\usepackage{booktabs}       
\usepackage{amsfonts}       
\usepackage{nicefrac}       
\usepackage{microtype}      
\usepackage{xcolor}         
\usepackage{hyperref}
\usepackage{url}
\usepackage{hyperref}
\usepackage{url}
\usepackage{bbm}
\usepackage{amsthm}
\newtheorem{theorem}{Theorem}
\newtheorem{corollary}{Corollary}
\usepackage{booktabs}
\usepackage{multirow}
\usepackage{graphics}
\usepackage{graphicx}
\usepackage{booktabs}
\usepackage{multirow}
\usepackage{caption}
\usepackage{etoolbox}
\usepackage{makecell}
\usepackage{listings}

\usepackage{hyperref}
\usepackage{url}
\usepackage{amsthm}
\usepackage{graphicx}
\usepackage{pifont}
\usepackage{bbm}
\usepackage{bm}
\usepackage{caption}
\usepackage{amsmath}
\usepackage{amssymb}
\usepackage{mathtools}

\theoremstyle{definition}

\usepackage{multirow}
\usepackage{adjustbox}
\usepackage[linesnumbered,ruled]{algorithm2e}
\usepackage{booktabs}
\usepackage{amssymb}
\usepackage{dirtytalk}
\usepackage{xurl}
\usepackage{thmtools}
\usepackage{thm-restate}
\usepackage{arydshln}
\usepackage{tablefootnote}
\usepackage{dirtytalk}
\usepackage{listings}
\usepackage{xcolor}

\title{Optimizing Input of Denoising Score Matching is Biased Towards Higher Score Norm}
\workshoptitle{Frontiers in Probabilistic Inference: Sampling Meets Learning}
\author{%
  Tongda Xu \\
  AIR \& CST, Tsinghua University\\
  \texttt{x.tongda@nyu.edu} \\
}

\begin{document}

\maketitle

\begin{abstract}
Many recent works utilize denoising score matching to optimize the conditional input of diffusion models. In this workshop paper, we demonstrate that such optimization breaks the equivalence between denoising score matching and exact score matching. Furthermore, we show that this bias leads to higher score norm. Additionally, we observe a similar bias when optimizing the data distribution using a pre-trained diffusion model. Finally, we discuss the wide range of works across different domains that are affected by this bias, including MAR for auto-regressive generation \citep{li2024autoregressive}, PerCo \citep{careil2023towards} for image compression, and DreamFusion \citep{poole2022dreamfusion} for text to 3D generation.
\end{abstract}

\section{Introduction}
The diffusion model is a type of generative model that consists of a Gaussian Markov chain in continuous space \citep{sohl2015deep,ho2020denoising,song2020score}. We denote the source image as $x_0$, and the forward diffusion process involves perturbing the image with Gaussian noise, i.e., $q(x_t) = x + \mathcal{N}(0, \sigma^2_t)$. To sample from the backward diffusion model, it is necessary to know the score function $\nabla_{x_t} \log q(x_t)$ in order to construct the reverse stochastic differential equation \citep{anderson1982reverse}.

The learning of diffusion models relies on the estimation of the score function $\nabla_{x_t}\log q(x_t)$. The most straightforward approach is explicit score matching (ESM), which employs a neural network $s_{\theta}(x_t, t)$ parameterized by $\theta$ to approximate the score:
\begin{gather}
   \mathcal{L}_{ESM}(\theta) = \mathbb{E}_{q(x_t|x)p(x)}[||s_{\theta}(x_t,t) - \nabla_{x_t} \log q(x_t)||^2].
\end{gather}
However, $\mathcal{L}_{ESM}$ is intractable because $\nabla{x_t}\log q(x_t)$ is itself intractable. \citep{vincent2011connection,Song2019GenerativeMB} proposed using denoising score matching (DSM) instead, as $\nabla_{x_t}\log q(x_t|x)$ is simply the score of a Gaussian distribution. Furthermore, \citep{vincent2011connection} proved the equivalence between the $\mathcal{L}_{ESM}$ objective and the $\mathcal{L}_{DSM}$ objective when optimizing $\theta$:
\begin{gather}
    \mathcal{L}_{DSM}(\theta) = \mathbb{E}_{q(x_t|x)p(x)}[||s_{\theta}(x_t,t) - \nabla_{x_t} \log q(x_t|x)||^2].
\end{gather}

\begin{theorem} \citep{vincent2011connection} When optimizing $\theta$, the denoising score matching is equivalent to explicit score matching, \textit{i.e.}, 
\begin{gather}
    \mathcal{L}_{DSM}(\theta) \smallsmile \mathcal{L}_{ESM}(\theta).
\end{gather}
\end{theorem}
Many recent works have begun to use $\mathcal{L}_{DSM}$ to optimize targets other than $\theta$. For example, MAR \citep{li2024autoregressive} adopts $\mathcal{L}_{DSM}$ to optimize $c$, the conditional input of the diffusion model to train an auto-regressive image generation model:
\begin{gather}
    \theta^*,c^* \leftarrow \arg\max \mathcal{L}_{DSM}(\theta,c) = \mathbb{E}_{q(x_t|x)p(x|c)}[\frac{1}{2}||s_{\theta}(x_t,t,c) - \nabla_{x_t}\log q(x_t|x,c)||^2].
\end{gather}
Another example is SDS \citep{poole2022dreamfusion} for text-to-3D generation, which utilizes a pre-trained diffusion model $s_{\theta}(.,.,.)$ and $\mathcal{L}_{DSM}$ to optimize the input distribution $p(x)$:
\begin{gather}
    p(x)^* \leftarrow \arg\max \mathcal{L}_{DSM}(p(x)) = \mathbb{E}_{q(x_t|x)p(x)}[\frac{1}{2}||s_{\theta}(x_t,t) - \nabla_{x_t}\log q(x_t|x)||^2].
\end{gather}
In this workshop paper, we show that the above usage of $\mathcal{L}_{DSM}$ is biased, as it breaks the equivalence between $\mathcal{L}_{DSM}$ and $\mathcal{L}_{ESM}$. Moreover, we demonstrate that this bias drives the optimization towards higher score norm. Finally, we discuss the broad range of works that are potentially affected by this bias.

\section{Optimizing Condition with Diffusion Loss is Biased Towards Higher Score Norm}
According to \citep{vincent2011connection}, $\mathcal{L}_{DSM}$ and $\mathcal{L}_{ESM}$ have the following relationship:
\begin{align}    
    \mathcal{L}_{DSM}(\theta,c) =& \mathcal{L}_{ESM}(\theta,c) - \underbrace{\mathbb{E}_{q(x_t|x)p(x|c)}[\frac{1}{2}||\nabla_{x_t}\log q(x_t|c)||^2]}_{C_2} \notag \\&+ \underbrace{\mathbb{E}_{q(x_t|x)p(x|c)}[\frac{1}{2}||\nabla_{x_t}\log q(x_t|x,c)||^2]}_{C_3}.
\end{align}
When optimizing $\theta$, the $C_2$ and $C_3$ terms cancel out, as they are not related to $\theta$. However, although $\theta$ is not related to these terms, the condition $c$ is involved in both the $C_2$ and $C_3$ terms. Since $x_t$-$x$-$c$ forms a Markov chain, the $C_3$ term has no gradient with respect to $c$.

Therefore, when optimizing $c$ using $\mathcal{L}_{DSM}$, there is an additional term compared to $\mathcal{L}_{ESM}$. Since the equivalence between $\mathcal{L}_{DSM}$ and $\mathcal{L}_{ESM}$ is broken, $\mathcal{L}_{DSM}$ becomes biased.

\begin{theorem} When optimizing both $\theta$ and $c$, the denoising score matching is equivalent to explicit score matching with a bias term, \textit{i.e.},
\begin{gather}
    \mathcal{L}_{DSM}(\theta,c) \smallsmile \mathcal{L}_{ESM}(\theta,c) - \underbrace{\mathbb{E}_{q(x_t|x)p(x|c)}[\frac{1}{2}||\nabla_{x_t}\log q(x_t|c)||^2]}_{C_2}.
\end{gather}
\end{theorem}
We have shown that for the condition $c$, minimizing $\mathcal{L}_{DSM}$ is equivalent to simultaneously minimizing $\mathcal{L}_{ESM}$ and maximizing $C_2 = \mathbb{E}{q(x_t|x) p(x|c)}\left[\frac{1}{2}||\nabla{x_t}\log q(x_t|c)||^2\right]$. And therefore, optimizing $\mathcal{L}_{DSM}$ is biased towards higher score norm.

\section{Optimizing Data Distribution with Diffusion Loss is Also Biased Towards Higher Score Norm}
A similar result can be trivially shown when using the score matching diffusion loss with a pre-trained diffusion model to optimize the input data distribution $p(x)$.
\begin{corollary}
When optimizing input distribution $p(x)$ with pre-trained diffusion model, the denosiing score matching is equivalent to explicit score matching with a bias term, \textit{i.e.,}
\begin{align}
    \mathcal{L}_{DSM}(p(x)) \smallsmile \mathcal{L}_{ESM}(p(x)) - \mathbb{E}_{q(x_t|x)p(x)}[\frac{1}{2}||\nabla_{x_t}\log q(x_t)||^2].
\end{align}
\end{corollary}
\section{Discussion: Works That are Effected by This Bias}
Optimizing the condition or input distribution using the diffusion loss is a common technique in the application of diffusion models across various domains. Below, we highlight four areas with recent works that are affected by this bias; this list is by no means exhaustive.

\textbf{Auto-regressive Generation}: Various works in tokenization-free auto-regressive generation are affected by this bias. MAR \citep{li2024autoregressive} adopts the diffusion loss to optimize $c$, which is the output of the auto-regressive model. MetaQuery \citep{pan2025transfer} employs diffusion loss to optimize $c$, which is the output of the connector in MLLM models. Similarly, \citep{deng2024autoregressive,chen2024diffusion,Yu2025FrequencyAI,Han2025NextStep1TA,Zhao2025DiSADS,Fan2024FluidSA,Tang2024HARTEV,Meng2024AutoregressiveSS} are also effected by this bias.

\textbf{Image Compression}: CDC \citep{yang2023lossy} jointly minimizes the bitrate of $c$ alongside the diffusion loss. PerCo \citep{careil2023towards} optimizes $c$ with the diffusion loss, where the resulting gradient is used to refine a VQ-VAE. FlowMo \citep{Sargent2025FlowTT} optimizes $c$ with diffusion loss, whose gradient is used to jointly train a VQ-VAE \citep{van2017neural}.

\textbf{Text-to-3D Generation}: DreamFusion \citep{poole2022dreamfusion} utilizes a pre-trained 2D diffusion model to optimize the input distribution $p(x)$ using the diffusion loss. The gradient with respect to $p(x)$ is then used to optimize a 3D representation model that is employed to render $p(x)$. Similarly, \citep{wang2024taming,Chung2024CFGMC} are also effected by this bias.

\textbf{Diffusion-based Inverse Problem Solving}: RED-Diff \citep{mardani2023variational} adopts the loss function used in DreamFusion to optimize $p(x)$, thereby solving inverse problems for operators such as super-resolution and de-blurring. Similarly, \citep{alkan2023variational,Pandey2025VariationalCF} are also effected by this bias.
\medskip
{
\small
\bibliographystyle{unsrt}
\bibliography{main}
}


\appendix

\section{Proof of Main Results}

\textbf{Theorem 2. }\textit{When optimizing both $\theta$ and $c$, the denoising score matching is equivalent to explicit score matching with a bias term, \textit{i.e.},
\begin{gather}
    \mathcal{L}_{DSM}(\theta,c) \smallsmile \mathcal{L}_{ESM}(\theta,c) - \underbrace{\mathbb{E}_{q(x_t|x)p(x|c)}[\frac{1}{2}||\nabla_{x_t}\log q(x_t|c)||^2]}_{C_2}.
\end{gather}
}
\begin{proof}
    \begin{align}
    \mathcal{L}_{ESM}(\theta,c)=&\mathbb{E}_{q(x_t|c)}[\frac{1}{2}||s_{\theta}(x_t,t,c) - \nabla_{x_t}\log q(x_t|c)||^2] \notag \\
    =& \mathbb{E}_{q(x_t|c)}[\frac{1}{2}||s_{\theta}(x_t,t,c)||^2] - \underbrace{\mathbb{E}_{q(x_t|c)}[s_{\theta}(x_t,t,c)^T\nabla_{x_t}\log q(x_t|c)]}_{S_1} \notag \\ &+ \mathbb{E}_{q(x_t|c)}[\frac{1}{2}||\nabla_{x_t}\log q(x_t|c)||^2]. \label{eq:esm}
    \end{align}
According to \citep{vincent2011connection}, the $S_1$ term can be transformed by applying log derivative twice:
    \begin{align}
        S_1 =& \mathbb{E}_{q(x_t|c)}[s_{\theta}(x_t,t,c)^T\nabla_{x_t}\log q(x_t|c)] \notag \\
        =&\int_{x_t}q(x_t|c) s_{\theta}(x_t,t,c)^T\nabla_{x_t}\log q(x_t|c) dx_t \notag \\
        =&\int_{x_t}q(x_t|c) s_{\theta}(x_t,t,c)^T\frac{\nabla_{x_t} q(x_t|c)}{q(x_t|c)} dx_t \notag \\
        =& \int_{x_t} s_{\theta}(x_t,t,c)^T\nabla_{x_t} q(x_t|c) dx_t \notag \\
        =& \int_{x_t} s_{\theta}(x_t,t,c)^T(\nabla_{x_t}\int_{x}p(x|c)q(x_t|x,c)dx) dx_t \notag \\
        =& \int_{x_t} s_{\theta}(x_t,t,c)^T(\int_{x}p(x|c)\nabla_{x_t}q(x_t|x,c)dx) dx_t \notag \\
        =& \int_{x_t} s_{\theta}(x_t,t,c)^T(\int_{x}p(x|c)q(x_t|x,c)\nabla_{x_t}\log q(x_t|x,c)dx) dx_t \notag \\
        =& \int_{x_t}\int_{x}p(x|c)q(x_t|x,c) s_{\theta}(x_t,t,c)^T(\nabla_{x_t}\log q(x_t|x,c)dx) dx_t \notag \\
        =& \mathbb{E}_{q(x_t|x)p(x|c)} [s_{\theta}(x_t,t,c)^T\nabla_{x_t}\log q(x_t|x,c)] \label{eq:s1}.
    \end{align}
Substituting Eq.~\ref{eq:s1} into Eq.~\ref{eq:esm}, we have
\begin{align}
    \mathcal{L}_{ESM}(\theta,c)=& \mathbb{E}_{q(x_t|c)}[\frac{1}{2}||s_{\theta}(x_t,t,c)||^2] - \mathbb{E}_{q(xt,x|c)} [s_{\theta}(x_t,t,c)^T\nabla_{x_t}\log q(x_t|x,c)] \\ \notag &+ \mathbb{E}_{q(x_t|c)}[\frac{1}{2}||\nabla_{x_t}\log q(x_t|c)||^2].
\end{align}
On the other hand, we have
\begin{align}
    \mathcal{L}_{DSM}(\theta,c) =& \mathbb{E}_{q(x_t|x)p(x|c)}[\frac{1}{2}||s_{\theta}(x_t,t,c) - \nabla_{x_t}\log q(x_t|x,c)||^2] \notag \\
    =& \mathbb{E}_{q(x_t|c)}[\frac{1}{2}||s_{\theta}(x_t,t,c)||^2] - \mathbb{E}_{q(xt,x|c)} [s_{\theta}(x_t,t,c)^T\nabla_{x_t}\log q(x_t|x,c)] \notag \\
    &+ \underbrace{\mathbb{E}_{q(x_t|x)p(x|c)}[\frac{1}{2}||\nabla_{x_t}\log q(x_t|x,c)||^2]}_{C_3} \notag \\
    =& \mathcal{L}_{ESM}(\theta,c) - \mathbb{E}_{q(x_t|c)}[\frac{1}{2}||\nabla_{x_t}\log q(x_t|c)||^2] + C_3.
\end{align}
For the $C_3$ term, we notice that $x_t-x-c$ forms a Markovian chain, and therefore 
\begin{align}
    C_3 = \mathbb{E}_{q(x_t|x)p(x|c)}[\frac{1}{2}||\nabla_{x_t}\log q(x_t|x)||^2].
\end{align}
As $C_3$ is not related to $\theta$ and $c$, we have
\begin{gather}
    \mathcal{L}_{ESM}(\theta,c) - \mathbb{E}_{q(x_t|c)}[\frac{1}{2}||\nabla_{x_t}\log q(x_t|c)||^2] \smallsmile \mathcal{L}_{DSM}(\theta,c).
\end{gather}
\end{proof}


\end{document}